\documentclass{article}
\usepackage[final,nonatbib]{neurips_2018}

\usepackage[utf8]{inputenc} 
\usepackage[T1]{fontenc}    
\usepackage{hyperref}       
\usepackage{url}            
\usepackage{booktabs}       
\usepackage{amsfonts}       
\usepackage{nicefrac}       
\usepackage{microtype}      
\usepackage{amsmath,amssymb}
\usepackage{algorithmic}
\usepackage{subcaption}
\usepackage{graphicx}
\usepackage{tabularx} 

\usepackage{multirow}
\usepackage[boxruled,linesnumbered]{algorithm2e}

\DeclareMathOperator*{\argmax}{arg\,max}

\newcommand{\MB}[1]{\mathbf{#1}}
\newcommand{\MBB}[1]{\mathbb{#1}}

\newcommand{\MC}[1]{\mathcal{#1}}

\newcommand{\1}{\MB{1}}

\newcommand{\smalleqb}[1]
{
	\begingroup
	\makeatletter
	\def
	\f@size{1}
	{#1}
    \endgroup
}  

\newtheorem{define}{Definition}

\newcommand{\ncl}{C}
\newcommand{\nlP}{\mathbf{O}}
\newcommand{\nlp}{\vec{o}}
\newcommand{\nlps}{o}

\newcommand{\nyw}[3]{\vec{x}_{#1}|_{#2}^{#3}}
\newcommand{\ny}{{\vec{x}}}
\newcommand{\nY}{\mathbf{X}}

\newcommand{\nH}{\mathbf{H}}

\newcommand{\nd}{d}

\newcommand{\nf}{{\vec{f}}}
\newcommand{\nff}[1]{{\vec{f}^{#1}}}
\newcommand{\nfx}{n_f}

\title{Deep-Aligned Convolutional Neural Network \\ for Skeleton-based Action Recognition 
	\\ and Segmentation}

\author{%
	Babak Hosseini
	\thanks{
		Preprint of the publication~\cite{hosseini2019deep} including extended experiments, as provided by the authors.
		The final publication is available at \url{https://ieeexplore.ieee.org/}
	} \\
	Center of Cognitive Interactive Technology (CITEC)\\
	Bielefeld University, Germany\\
	\texttt{bhosseini@techfak.uni-bielefeld.de}	
	\And
	Romain Montagne\\
	Eurodecision, 
	Versailles, France\\
	\texttt{r.montagne@hotmail.fr} \\
	\And
	Barbara Hammer\\
	Center of Cognitive Interactive Technology (CITEC)\\
	Bielefeld University, Germany\\
	\texttt{bhammer@techfak.uni-bielefeld.de} \\
}

\begin{document}
\maketitle

\begin{abstract}
	Convolutional neural networks (CNNs) are deep learning frameworks which are well-known for their notable performance in classification tasks. 
	%
	Hence, many skeleton-based action recognition
	and segmentation 
	(SBARS) 
	algorithms benefit from them in their designs.
	However, a shortcoming of such applications is the general lack of spatial relationships between the input features in such data types.
	Besides, non-uniform temporal scalings is a common issue in skeleton-based data streams which leads to having different input sizes even within one specific action category. 
	In this work, we propose a novel deep-aligned convolutional neural network (DACNN) to tackle the above challenges for the particular problem of SBARS.    
	Our network is designed by introducing a new type of filters in the context of CNNs which are trained based on their alignments to the local subsequences in the inputs. 
	These filters result in efficient predictions as well as learning interpretable patterns in the data.
	%
	%
	We empirically evaluate our framework on real-world benchmarks showing that the proposed DACNN algorithm obtains a competitive performance compared to the state-of-the-art while benefiting from a less complicated yet more interpretable model.	
\end{abstract}

\section{Introduction}\label{sec:intro}
Action recognition is still a challenging problem in the area of computer vision despite its numerous applications in sports analysis, video surveillance, and human-machine interaction~\cite{aggarwal2011human,poppe2010survey}.
A popular way to represent action data is through its description based on 3D skeleton sequences, which are the 3D positions of body joints through time. 
This representation medium can efficiently describe the underlying dynamics of the actions~\cite{johansson1973visual}.
Recently, 
the development of motion capture technologies such as the Microsoft Kinect~\cite{zhang2012microsoft} 
along with the existing robust pose estimation methods~\cite{shotton2011real}
has considerably facilitated the collection of skeleton-based datasets. 
These technological advancements have motivated considerable research efforts being put into this area~\cite{aggarwal2014human,du2015hierarchical}.

In particular, skeleton-based action recognition has become an interesting problem for many deep learning algorithms such as convolutional neural networks (CNNs)~\cite{du2015skeleton,ke2017new,yan2018spatial} and recurrent neural networks (RNN)~\cite{du2015hierarchical,song2017end,liu2018skeleton}.
RNN methods can learn the temporal dynamics of the sequential data; nevertheless, they have practical shortcomings in the training of their stacked structures~\cite{liu2017enhanced,wang2017modeling}.

Compared to RNN architectures, CNN-based methods provide more effective solutions by extracting local features from their input and finding discriminative patterns in the data~\cite{gehring2017convolutional,ke2017new}. 
However, regardless of the promising feature extraction capability of CNN, its specific convolutional structure is designed originally for image-based input data and primarily relies on spatial dependencies between the neighboring points.
In contrast, such a direct relationship does not generally exist in skeleton-based action datasets. 
Although some works tried to solve this problem by using 1-dimensional filters (only for the temporal dimension), it is still not an efficient solution to this specific shortcoming of CNN-based frameworks~\cite{zheng2014time}.

A crucial step before analyzing any motion data is the temporal segmentation phase, 
in which we 
predict the action to which each time-frame belongs.
Although there exist unsupervised algorithms for temporal segmentation of motion data~\cite{kruger2017efficient,zhou2013hierarchical}, 
they usually oversegment actions into smaller sub-actions
or segment also the blank parts of the stream.
%

\textbf{Motivations:}
%
An essential group of techniques for classification of the sequential data is time-series alignment methods~\cite{petitjean2016faster}.
It is shown that via comparing each data sequence to some predefined or learned sequences, we can discriminate or segment the data samples with high accuracy ~\cite{anagnostopoulos2006global,rakthanmanon2013data}.
Also, in algorithms similar to~\cite{ye2009time}, finding a small distinct subsequence in the input data can reveal its classification label. 
Therefore, it is logical to design filters in CNN architecture based on the alignment concept to extract essential features in sequential data.



\begin{figure}
	\centering
	\includegraphics[width=.81\linewidth]{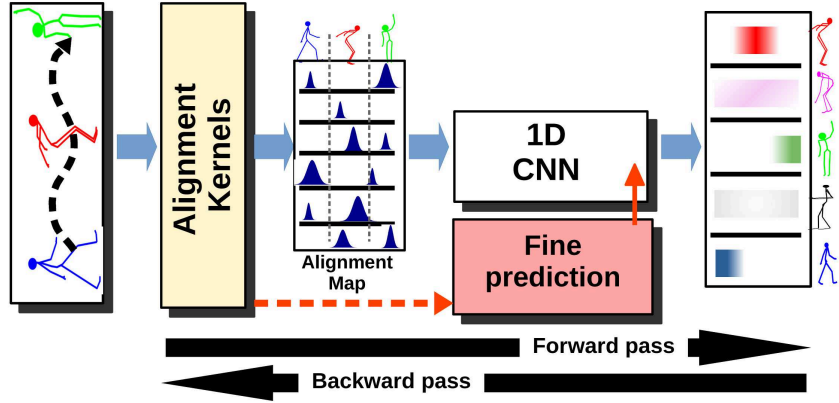}
	\caption{General overview of DACNN framework.
		Alignment kernels preprocess the input streams. 
		1D-CNN performs temporal prediction based on the derived alignment map.
		The fine-prediction unit improves the resolution of output prediction.}
	\label{fig:alg}
\end{figure}

\textbf{Contributions:}
In this work, we propose a novel framework named deep-aligned convolutional neural network (DACNN) as an efficient skeleton-based action recognition
 and segmentation 
 (SBARS) algorithm.
As depicted in Fig.~\ref{fig:alg}, DACNN is constructed upon the activation maps of the alignment kernels in its first layer following by a CNN with 1-dimensional convolutional filters (1D-CNN). 
This architecture makes DACNN flexible to the input size. 
%

To be more specific,
	We introduce the alignment kernels (Al-filters) 
	in the context of CNN, which are more efficient than the convolutional filters
	regarding the temporal feature extraction and classification of skeleton-based action data.	
	DACNN learns temporal subsequences in the data as essential local patterns which make the outcome of the network semantically interpretable
	and leads to more accurate predictions. 
	

In the next section, we summarize the most relevant work in the literature. 
In Sec.~\ref{sec:al-fil}, we introduce the alignment filters while we explain the architecture of DACNN in Sec.~\ref{sec:DACNN}. 
The empirical evaluation of DACNN is provided in Sec.~\ref{sec:exp}, and we conclude the paper in the last section.

\section{Related Work}
Generally, it is possible to split the skeleton-based action recognition methods into two general categories:

The first group includes methods with a pre-processing step to extract features (usually hand-coded) which best represent the skeleton information.
For instance, in~\cite{wang2012mining} the local occupancy pattern was proposed based on the depth appearance of the joints along with an ensemble action recognition model.
In~\cite{hussein2013human}, they proposed 
a discriminator based on the covariance matrix of joints locations, 
while in~\cite{vemulapalli2014human}, the algorithm was designed based on
the 3D geometric relationships between different regions of the body.

The second category contains deep learning feature extraction methods which are generally designed based on CNN and RNN models.
Among RNN frameworks, a regularized LSTM architecture is proposed in~\cite{zhu2016co} for co-occurrence feature extraction.
A spatiotemporal attention-based model is utilized in~\cite{song2017end} to assign different weights to different frames, and a trust-gate technique was proposed in~\cite{liu2018skeleton} to deal with the noise in skeleton-based data.

Regarding the approaches based on CNN models, Tang \textit{et al.}~\cite{tang2018deep} combined CNN with a reinforcement learning module to learn the most efficient video frames. 
In~\cite{ke2017new}, cylindrical coordinates were utilized to present a new skeleton representation. The skeleton data was transformed into images in~\cite{liu2017enhanced} to be more appropriate for CNN architecture,
and Li \textit{et al.}~\cite{wang2017modeling} combined two CNN models assigned to the joint position and velocity information individually.
\section{Alignment Kernels for CNN}\label{sec:al-fil}
\begin{figure}
	\begin{center}
		\includegraphics[width=1\linewidth]{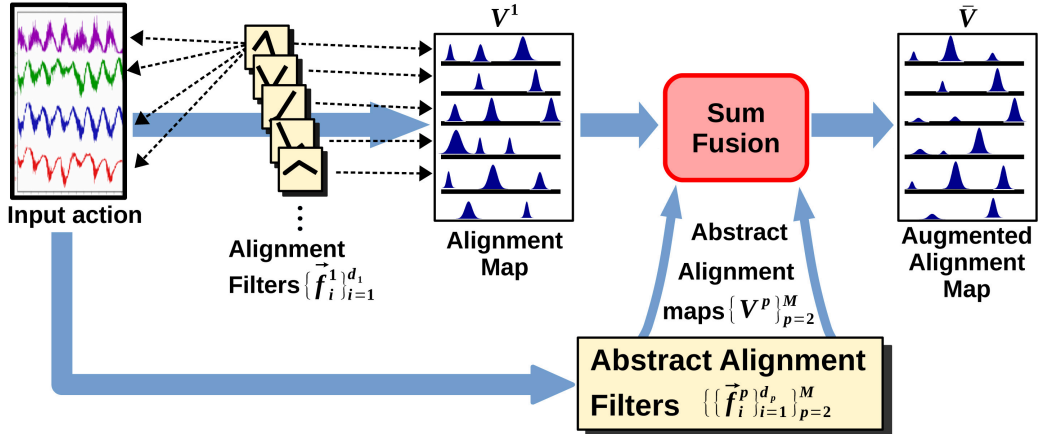}
	\end{center}
	\caption{The alignment layer of the network. 
		Each Al-filter $\nf_i^1$ is applied to all $\nfx$ channels in the input to form the $i$-th row in the alignment map $\MB{V}^1$.
		Adding $\MB{V}^1$ to the alignment maps of other Abs-filters derives the augmented map $\bar{\MB{V}}$.}
	\label{fig:al-fil}
\end{figure}

Principally, the input layer of a CNN consists of $\nfx$ different channels $\{\ny_j\}_{j=1}^{\nfx} \in \MBB{R}^T$ each of which contains the temporal skeleton data related to one dimension of $\nY$.
Based on the discussed rationale in the previous section, we propose the following distance-based non-linear kernel as the fundamental feature extraction unit of our DACNN architecture (Alignment layer in Fig.~\ref{fig:alg}):
\begin{equation}
g(\nyw{j}{t_0}{t},\nf_i^1)=e^{-\|\nf_i^1-\nyw{j}{t_0}{t}\|_2^2},~~\forall i=1,\dots,d_1,
\label{eq:filt}
\end{equation}
where $\{\nf_i^1\}_{i=1}^{d_1} \in \MBB{R}^t$ are the alignment filters with the receptive field of $t$, and $\nyw{j}{t_0}{t}$ denotes any temporal window of length $t$ starting from the $t_0$-th frame of a $\ny_j$ channel.
After scanning each channel of $\nY$ by these filters with a stride $s$ (Fig.~\ref{fig:al-fil}), we obtain a tensor $\MB{V}\in \MBB{R}^{T\times \nfx \times d_1}$ as the activation map. 
Each entry $v_{jik}$ from $\MB{V}$ represents the similarity between the $j$-th window in channel $\ny_i$ and the filter $\nf_k$ and summing $\MB{V}$ over its second dimension results in the more summarized activation map $\MB{V}^1 \in \MBB{R}^{\nd_1 \times T}$ (Fig.~\ref{fig:al-fil}).

To make an analogy to a regular CNN structure, 
we reformulate the introduced alignments 
as an $l_2$-norm comparison layer followed by an activation-layer of $f(x)=e^{-x}$ units. 
This reformulation is conceptually similar to the convolution and RELU layers of a regular CNN architecture.
Therefore, a proper classification-based training scenario leads to finding discriminative patterns in $\MB{V}^1$ with high activation values (near $1$). 
In other words, we learn filters $\nf_i$ which can distinguish the data classes based on their similarities to the local parts in the data channels $\{\ny_j\}_{j=1}^{\nfx}$.

\subsection{Vanishing Gradient and Saturated Activation}\label{sec:vanish}
Updating the values of an Al-filters $\nf$ in a gradient-based optimization routine, 
the activation function of (\ref{eq:filt}) and its gradient obtain infinitesimal values if $\|\nf-\nyw{j}{t_0}{t}\|_2^2$ becomes large. 
This condition leads to zero updates of $\nf_i$ in the optimization process
which may occur when a filter $\nf_i$ has a large distance to
all subsequences in $\nY$ (e.g., lousy initialization) or the learning rate is too high.

As a systematic workaround, we compute the activation map $\MB{V}^1$ using the function
\begin{equation}
g(\vec x,\vec f)=(1+a)e^{-\|\vec x-\vec f\|_2^2}-a
\label{eq:filt2}
\end{equation}
instead of (\ref{eq:filt}) by choosing a small $a$ (in implementations we use $a=0.1$).
The value of $g(\vec x,\vec f)$ becomes $-a$ when $|\vec x-\vec f|$ is large, 
we preserve the sparseness effect of the activation function when computing $\MB{V}^1$ 
and leads to faster convergence.
In addition, in the backward phase of the training, we employ the conditional gradient
\begin{equation}
\nabla_{\vec{f}} g=
\begin{cases}
	-2(1+a)e^{-\|\vec{\Delta}\|_2^2}\vec{\Delta} & \|\vec{\Delta}\|_2^2\le 0.5\\
	-2[(1+a)e^{-\|\vec{\Delta}\|_2^2}+1]\vec{\Delta}+2\sqrt{0.5}sign(\vec{\Delta}) & \|\vec{\Delta}\|_2^2 > 0.5
\end{cases} 
\label{eq:grad}
\end{equation}
where $\vec{\Delta}=\vec x-\vec f$, and $\sqrt{0.5}$ is the point from which the gradient of (\ref{eq:filt}) starts to decrease toward zero.
In such case, the gradient becomes $-2(\nyw{j}{t_0}{t}-\nf)$ 
when $\|\nf-\nyw{j}{t_0}{t}\|_2^2$ has a relatively large value and prevents the filter weights from becoming saturated. 
%
%
More detail explanation of the above is provided in the Appendix section.
\subsection{Abstract Filters}
\begin{figure}[!b]
	\begin{center}
		\includegraphics[width=1\linewidth,height=3cm]{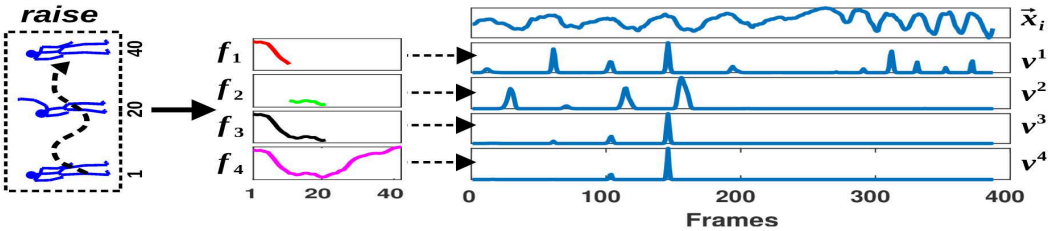}
	\end{center}
	\caption{Abstract filters $\{\nf_4,\nf_3\}$ result in sparse alignment maps $\{v^4,v^3\}$, and
		$\nf_4$ can be interpreted as the representative of a complete arm movement in the \textit{raise} action.}
	\label{fig:abs-ex}
\end{figure}

Although the proper training of the Al-filters can fit them to the small local patterns in $\nY$ which are essential to the classification task, we are also interested in finding longer patterns in the action data.
The benefits of finding these patterns are two folds: 
\\1. Applying longer filters on the data leads to sparser activation maps (Fig.~\ref{fig:abs-ex}: $\{v^4,v^3\}$ vs. $\{v^1,v^2\}$).
\\2. Long patterns are semantically more interpretable than the short ones
(Fig.~\ref{fig:abs-ex}: $v^4$ represents a complete action).

To that aim, we define the abstract filters (Abs-filter) $\{\nf^p_i\}_{i=1}^{\nd_p}$ with the receptive field of $pt$, where $p \in \MBB{N}$.
Each $\nf^p_i$ is a temporal concatenation of $p$ smaller Al-filter of size $t$ as 
$
\nff{p}_i=\underset{j\in I}{\oplus} \nf^1_j,
$
where $\underset{j\in I}{\oplus}$ concatenates the Al-filters $\nf^1_j$ with the index order given by the set $I$.
To find $\nff{p}_i$ filters systematically, we select the potential candidates among the Al-filters as the first step.
\begin{define}\label{def:abs}
	Two Al-filters $\{\nff{1}_i,\nff{1}_j\}$ of size $t$ are candidates to form an Abs-filter if 
	we can find a window of size $t$ starting at the time-frame $t_0$ of a data channel $\ny_k$ such that 
	\begin{equation}
	\rho(\nff{1}_i,\nff{1}_j,\nyw{k}{t_0}{t})=g(\nyw{k}{t_0}{t},\nff{1}_i)g(\nyw{k}{t_0+t}{2t},\nff{1}_j) \ge r,
	\label{eq:thresh}
	\end{equation} 	
	where $r$ is a scaler with a value sufficiently close to 1.
\end{define}
Based on the definition~\ref{def:abs}, 
when $\rho(\nf^1_i,\nf^1_j,\nyw{k}{t_0}{t})$ has a value close to $1$, 
there exists a temporal pattern in a channel of $\nY$ which fits the concatenation of $(\nf^1_i,\nf^1_j)$.
Accordingly, by using a moderate threshold in ~(\ref{eq:thresh}) (e.g., $r=0.8$), 
we collect all the candidate Al-filters $\nf^1_i$ in the forward pass (Fig.~\ref{fig:alg}) and form a binary-weighted graph $\MC{G}$. 
In this graph, the filters $\{\nf^1_i\}_{i=1}^{d_1}$ are the nodes, those of which correspond to the definition~\ref{def:abs} have undirected links of weight $-1$ between them.
Now, we find the existing Abs-filters of different sizes via finding shortest paths between connected nodes of $\MC{G}$, which is efficiently solved by the Floyd Warshall algorithm \cite{hougardy2010floyd}. 
So, those Abs-filters which are subsets of the longer ones are detected and eliminated. 

After finding $\MB{M}$ Abs-filter, for each $p\le\MB{M}$, 
we apply the created filters $\{\nf^p_i\}_{i=1}^{d_p}$ on the channels of $\nY$ (analogous to Al-filters).
Then, the generated activation maps $\{\MB{V}^p\}_{p=2}^\MB{M}$ are added to $\MB{V}^1$ to result in the augmented map $\bar{\MB{V}}$ (Fig.~\ref{fig:al-fil}).
An immediate benefit of these long filters is the sparse alignments we obtain in each row of $\tilde{\MB{V}}$ (Fig.~\ref{fig:al-fil}), which enriched the feature extraction part of our framework (Fig.~\ref{fig:alg}) and increases the interpretability of the outcome (Fig.~\ref{fig:abs-ex}).
\section{Deep-Aligned CNN }\label{sec:DACNN}
In Sec.~\ref{sec:al-fil}, we introduced the alignment kernels as the important feature extraction layer of our skeleton-based action recognition algorithm (Fig.~\ref{fig:alg}).
Now, we discuss the role and rationale of the remaining parts in the DACNN architecture.

\begin{figure}[!b]
	\begin{center}
		\includegraphics[width=.81\linewidth]{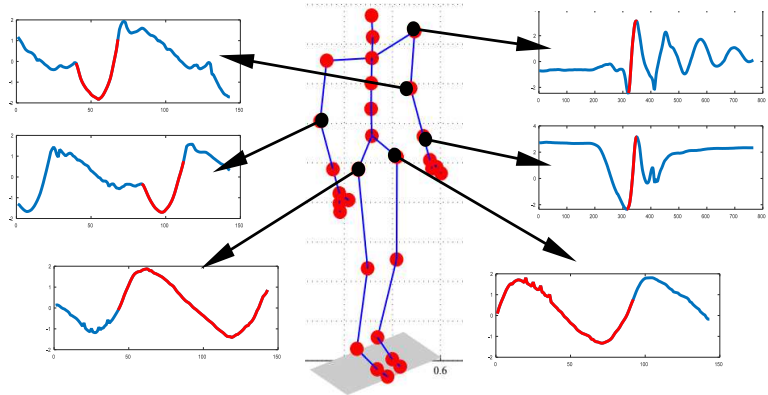}
	\end{center}
	\caption{Examples of similar subsequences (red curves) which are found among different body joints and in different temporal locations related to a skeleton-based action.}
	\label{fig:joint}
\end{figure}

For each real-life skeleton-based motion data $\nY$,
different data channel (dimension) contain streams of continues changes in the values of different joint orientations. 
These values particularly lay in the range of $[0~ 2\pi]$ throughout normalization (or in $[\theta_0~2\pi]$ due to physical limitations). 
Therefore, it is highly expected to find short subsequences in different dimensions and temporal locations in $\nY$ (or long patterns in symmetrical joints), which have similar shapes or curvatures (Fig.~\ref{fig:joint}).
A similar characteristic can also be observed in the quaternion representation of the $\nY$.

Based on this observation, we can extract the important patterns in dimension of $\nY$ by applying each filter $\nf^p_i$ to all the channels. 
As a direct benefit, this structure notably reduces down the network's number of parameters and avoids an unnecessary model complexity.

After deriving the augmented activation map $\bar{\MB{V}}$ (Fig.~\ref{fig:al-fil}), we feed it to a regular CNN which contains 1-dimensional convolution filters (1D-conv) as in Fig.~\ref{fig:CNN}. Each deep layer $q$ of the network contains two 1D-conv layers following a max-pooling layer with the stride of $2$. 
This combination results 
in a feature map $l_q$ for each deep layer $q$ with the temporal size of $\MBB{R}^{T/2^q}$.
Hence, the data representation becomes more abstract as we go through these layers.

As a complementary choice to Sec.~\ref{sec:vanish}, we apply the ELU operator \cite{clevert2015fast} to the input of each  max-pooling layer as
\begin{equation}
\omega(x)=
\frac{1}{2}[x+e^{x}-1+sign(x)(x-e^{x}+1)],
\end{equation}  
while $\frac{d\omega}{dx}|_{x=0}=1$ lets the gradients flow through the layers of 1D-CNN even if they belong to any saturated filter.
\subsection{Prediction Layer}
\begin{figure}[!t]
	\begin{center}
		\includegraphics[width=1\linewidth]{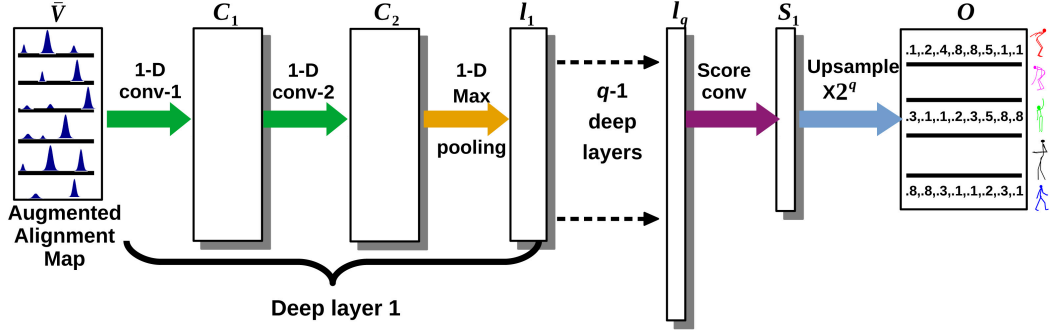}
	\end{center}
	\caption{The architecture of the 1D-CNN unit in the DACNN framework. It maps the augmented alignment map $\bar{\MB{V}}$ (input) to the prediction map $\MB{O}$ (output ).}
	\label{fig:CNN}
\end{figure}

We want our network structure to take inputs of different sizes 
and also to fit both the segmentation and classification problems.
To that aim, we design a convolutional prediction layer (inspired by \cite{long2015fully}) to obtain a prediction map $\nlP \in \MBB{R}^{T \times \ncl}$ as the output of 1D-CNN (last layer in Fig.~\ref{fig:CNN}).
Each entry $\nlps_{ct}$ represents the confidence of the network in assigning the $t$-th time-frame of $\nY$ to the class $c$.

We apply a convolution layer of size $1 \times \ncl$ (score-conv) to the last activation map ($l_q$ in Fig.~\ref{fig:CNN}) to map the abstract features to a score-map of size $\ncl \times \frac{T}{2^q}$ followed by a $\times 2^q$ upsampling. 
The upsampling is performed using the transposed convolution of size $2^q$ (A.K.A deconvolution) ~\cite{dumoulin2016guide}.
We compare the obtained prediction map $\MB{O}$ to the target label matrix $\nH$ using a cross-entropies loss function $\MB{LOSS}$ to calculate the prediction error (cost) of the network
\begin{equation}
\MB{LOSS}=-\sum_{c=1}^{C} \sum_{t=1}^T h_{ct} log(\nlps_{ct}).
\label{eq:loss}
\end{equation}

Therefore, 
after DACNN is converged to an optimal point, 
we can predict the label of each individual time-frame $t$ in $\nY$ as
$
	c=\underset{c}{\arg\max}~\|1-\nlps_{c t}\|_2^2.
$
%
However, in a classification setting where the whole $\nY$ sequence belongs to one class, we identify its class label as
$
c=\underset{c}{\arg\min}~\|\vec{\MB{1}}_{1\times T}-\nlp_{c,:}\|_2^2,		
$
in which $\nlp_{c,:}$ denotes the $c$-th row of $\nlP$.
Nevertheless, 
the training phase of DACNN is the same for both segmentation and classification tasks using the defined cost term $\MB{LOSS}$ in ~(\ref{eq:loss}).
The complete scheme of the DACNN framework while putting all units together is presented in the online supplementary material\footnote{https://github.com/bab-git/DACNN}.

\subsection{Fine-Prediction Module}
According to the structure of 1D-CNN, each pooling layer downsamples the feature maps by a factor of 2. 
Hence, for a network with $q$ deep layers, the prediction map $\nlP$ has the resolution of $2^q$ time-frames,
which reduces the accuracy of the network, especially in the segment borders.
Therefore, 
to have a fine-grained prediction map, we employ a specific skip-connection structure throughout the fine-prediction module in our DACNN framework (Fig.~\ref{fig:alg}). 

\begin{figure}[!t]
	\begin{center}
		\includegraphics[width=.81\linewidth]{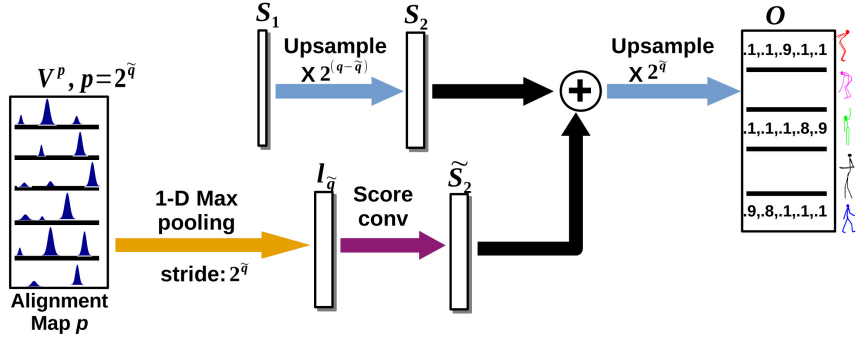}
	\end{center}
	\caption{Fine-prediction module of DACNN. The skip-connections from score-maps of specific Abs-filters result in fine-grained segmentations compared to Figure~\ref{fig:CNN}.}
	\label{fig:fine}
\end{figure}

As illustrated in Fig.~\ref{fig:fine}, 
we add the score maps obtained from the Abs-filters (from the alignment layer) to the upsampling process of 1D-CNN. 
More precisely,
for the Abs-filters of size $p=2^{\tilde{q}}t,\forall \tilde{q}\ge2$,
we downsample their alignment map $\MB{V}^p$ by a max-pooling of size $p$ and feed it to a score-conv layer.
This operation results in a score map $\tilde{S}_2 \in \MBB{R}^{\ncl\times \frac{T}{2^{\tilde{q}}}}$ that we add to the upsampled score map of a $q$-layer 1D-CNN (Fig.~\ref{fig:fine}).

As the rationale, each $\MB{V}^p$ has a sparse activation-map with gap intervals of $pt$ between the extracted matching patterns especially when $p\ge 2$ (e.g., Fig.~\ref{fig:abs-ex}). 
Hence, $p$-factor downsampling preserves the essential information existing in $\MB{V}^p$ while it still increases the resolution of $\MB{O}$ by order of $2^{(q-\tilde{q})}$ when $\tilde{q}<q$.
Additionally, this skip-connection module intensifies the role of the Al-filters in the prediction task due to the weight-sharing between them and the Abs-filter, which increases their discriminative quality.
\section{Experiments}\label{sec:exp}
%
%
%
In this section, we implement our DACNN framework on 3 well-known action recognition datasets in order to obtain empirical evaluation of its performance.\footnote{More implementation detail, additional experiments, and the code of DACNN framework are available from https://github.com/bab-git/DACNN}. 
\subsection{Datasets}
\textbf{CMU Mocap}~\cite{CMU_mocap}: 
This human motion capture dataset is a an extensive collection of different human activities which are recorded by the Vicon system. 
The data is collected from 144 subjects each of which performs a sequence of different actions in separate sessions. 
The segmentation is performed on each session and is evaluated based on the Jaccard index measure.
We collect the sessions which contain actions from 15 highly observed categories in the data including 
\textit{walk}, \textit{punch}, \textit{wave}, \textit{run}, \textit{jump}, and \textit{raise}, and consider the rest of actions as the blank spaces (gaps). The training set contains $75\%$ of the sessions, while the rest is left for testing. 

\textbf{Montalbano V2 dataset}~\cite{escalera2014chalearn}
This dataset is related to the "ChaLearn Looking at People" challenge which is recorded with Kinect technology as described by~\cite{escalera2014chalearn}.   
It includes 13,858 samples of 20 different Italian sign gestures which are recorded in continues sequences. The task of this challenge is to perform SBARS for sign gestures based on the given test and training streams. 
Following~\cite{escalera2014chalearn}, we employ the Jaccard index for performance evaluations. 

\textbf{SYSU-3D Human-Object Interaction dataset (SYSU)}~\cite{hu2015jointly}:
This dataset contains 12 different action classes recorded in 480 video sequences.
SYSU dataset is captured from 40 human subjects, and in each time-frame, it represents the 3D coordinates of 20 body joints.
We create the test and train batches by splitting the 
number of subjects in half over 10 randomized splits.

\textbf{NTU-RGB+D Dataset (NTU)}~\cite{shahroudy2016ntu}: 
This action recognition dataset consists of 60 classes of actions captured from 40 human subjects.
It has 56,000 sequences with 4 million frames in total, and the recorded data of 25 main body joints are used for the action recognition task.
There are two typical evaluation protocols for this benchmark:
The Cross-Subject (CS) recognition task which uses data of 20 subjects for training and the rest for testing,
and the Cross-View (CV) task in which the recorded samples from camera 2 and 3 constitute the training set and the rest is preserved for the test set.

\subsection{Implementation Details}
%
%
In the implementation of DACNN, we choose the kernel size of $t=3$ for Al-filters and $1\times 3$ for the conv-filters.
This choice of $t$ for Al-filters provides them enough angular freedom to learn any local curvature in the data.  
We also apply the dropout~\cite{srivastava2014dropout} of rate 0.65 to prevent DACNN from overfitting.
We train DACNN using the Adam approach~\cite{kingma2014adam}, and the skeleton data is normalized per dimension 
which allows the filters to be aligned to local parts of different joints.  
Regarding the implementation of the baseline algorithms, either we use their publicly available codes and tune their parameters with cross-validation on training set, or we refer to their reported results in the relevant publications. 
For each dataset, we use its typical evaluation setting reported in the literature.
%

\subsection{Action Segmentation Results}\label{sec:seg}
\begin{figure}
	\centering
	\includegraphics[width=.8\linewidth]{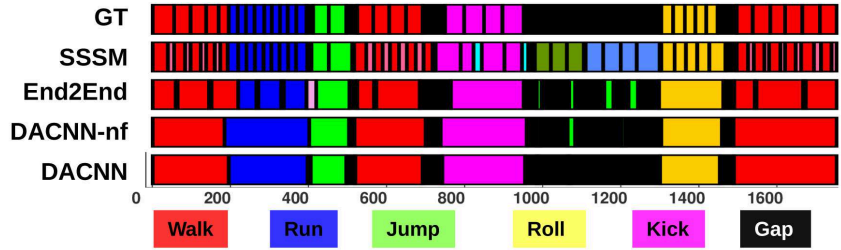}
	\caption{The segmentation results on the CMU dataset (Subject 86, trial 03). GT: ground truth segments.
		DACNN-nf: the DACNN framework without the fine-prediction unit (colored figure).}
	\label{fig:seg}
\end{figure}

\textbf{CMU Mocap:}
The segmentation performance of DACNN on CMU Mocap dataset is evaluated by its comparison to 
SSSM~\cite{kruger2017efficient} and HACA~\cite{zhou2013hierarchical} as unsupervised temporal segmentation approaches, 
and to 
ConvS2S~\cite{gehring2017convolutional},
End2End~\cite{ma2016end},
LSTM-CRF~\cite{lample2016neural},
and RNN-CRF~\cite{yang2016multi} as the deep sequence labeling frameworks.
In the latter group, we replace the word-embedding parts with the quaternion values of the time-frames. 
As reported in Table~\ref{tab:Mont}, the performances of deep learning algorithms have substantial distances from the 
unsupervised methods. 
Compared to them, DACNN outperforms the best method (End2End) with a notable margin of $4.1\%$. 
This result supports the effectiveness of the DACNN architecture regarding the supervised segmentation of skeleton-based action data.
\begin{table}[b]
	\centering	
	\footnotesize		
	\begin{tabular}{lc|lc} 
		\toprule
		\multicolumn{2}{c}{Montalbano}
		& \multicolumn{2}{|c}{CMU Mocap}\\
		\toprule
		Method   & \textit{JI} & Method   & \textit{JI}\\
		\midrule
		YNL~\cite{escalera2014chalearn}~(2015)&	27.1 &  HACA~\cite{zhou2013hierarchical}~(2013) & 71.4\\
		Terrier~\cite{escalera2014chalearn}~(2015)&	53.9&SSSM~\cite{kruger2017efficient}~(2017)&75.7\\
		Quads~\cite{escalera2014chalearn}~(2015)&	74.6&LSTM-CRF~\cite{lample2016neural}(2016)&85.9\\
		Ismar~\cite{escalera2014chalearn}~(2015)&	74.7&End2End~\cite{ma2016end}~(2016)&88.4\\
		Gesture Labeling~\cite{chang2014nonparametric}~(2014)&	78.4&RNN-CRF~\cite{yang2016multi}~(2016)&88.1\\
		CNN+LSTM2~\cite{nunez2018convolutional}~(2018)&79.5&ConvS2S~\cite{gehring2017convolutional}~(2017)&86.4\\
		Moddrop~\cite{neverova2016moddrop}~(2016)&	83.3&&\\
		\midrule
		DACNN (\textbf{Proposed})	&\textbf{87.2}&DACNN (\textbf{Proposed})&\textbf{92.5}\\				
		\bottomrule
	\end{tabular}
	\caption{Segmentation accuracy for Montalbano V2 and CMU Mocap based on Jaccard Index (\textit{JI}).}	
	\label{tab:Mont}
\end{table}
In Fig.~\ref{fig:seg}, the segmentation results of some baselines are visually evaluated on one of the challenging sequences from the CMU dataset (Subject 86, trial 03). 
%
In comparison to the unsupervised method SSSM, 
the supervised algorithms better identify the gaps (insignificant actions) because they are optimized by the relevant label information in the training phase.
Additionally, SSSM finds sub-clusters in some of the segments (e.g., \textit{walk} and \textit{kick}) which is not the desired outcome in supervised temporal segmentation.
%

The algorithms DACNN and DACNN-nf (without fine-prediction unit) have fewer mistakes in the results, especially in the gap areas.
Besides, DACNN obtains less overlapping between the segments 
and better predictions in the gap areas compared to DACNN-nf,
which emphasizes the positive effect of the fine-prediction module in DACNN.
%
%
In comparison, End2End has lower performance than DACNN and DACNN-nf regarding the gap areas and segments.

\textbf{Montalbano V2 dataset:}
For this dataset, we compare DACNN to the following baselines regarding the temporal segmentation accuracy:
YNL~\cite{escalera2014chalearn},
Terrier~\cite{escalera2014chalearn},
Quads~\cite{escalera2014chalearn},
Ismar~\cite{escalera2014chalearn},
Gesture Labeling~\cite{chang2014nonparametric},
Moddrop~\cite{neverova2016moddrop}, and
CNN+LSTM2~\cite{nunez2018convolutional}.
The majority of these algorithms
need to perform a separate segmentation on the data prior to the training phase, but the DACNN's training is independent of that preprocessing phase.
According to Table~\ref{tab:Mont}, our proposed method obtained higher performance than the baselines ($2.9 \%$ margin in the Jaccard index),
while the best alternative approach (Moddrop) has a relevantly complex network architecture and uses 3 different modalities (video, Mocap, and audio).
%
%

\subsection{Action Recognition Results}\label{sec:act}
\textbf{SYSU dataset:}
For the empirical evaluations, we compare our DACNN algorithm to other baselines including 
CNN+DPRL~\cite{tang2018deep},
ST-LSTM+Trust Gate~\cite{liu2018skeleton},
Dynamic Skeletons~\cite{hu2015jointly},
LAFF(SKL)~\cite{hu2016real},
SR-TSL~\cite{si2018skeleton},
VA-LSTM~\cite{zhang2017view}, and
GCA-LSTM~\cite{liu2017global}, 
which includes the most recent deep learning applications (CNN, LSTM, etc.) on this dataset.
In Table~\ref{tab:SYS}, our proposed DACNN framework outperforms the state-of-the-art with a $2.3 \%$ margin. 

\begin{table}[!b]
	\centering	
			\footnotesize	
	\begin{center}
		\begin{tabularx}{0.7\textwidth}{Xc|Xc} 
			\toprule				
			Method   & Acc. & Method   & Acc. \\
			\midrule		
			LAFF(SKL)~\cite{hu2016real}~(2016)&	55.2&CNN+DPRL~\cite{tang2018deep} (2018) &   76.7\\
			Dynamic Sk.~\cite{hu2015jointly}~(2015)&	75.2&GCA-LSTM~\cite{liu2017global}~(2017)&	78.6\\
			ST-LSTM+TG~\cite{liu2018skeleton} (2016) &  76.7&VA-LSTM~\cite{zhang2017view}~(2017)	&77.8\\
			SR-TSL~\cite{si2018skeleton}~(2018)	&82.0&DACNN (\textbf{Proposed})   &\textbf{84.3}\\						
			\bottomrule				 			
		\end{tabularx}
\end{center}
\caption{Recognition accuracy ($\%$) for SYSU-3D dataset.}
\label{tab:SYS}
\end{table}

\textbf{NTU dataset:}
%
For the NTU dataset, we evaluate DACNN in comparison to the state-of-the-art methods  
from the literature:
HBRNN-L~\cite{du2015hierarchical},
Dynamic Skeletons~\cite{hu2015jointly},
LieNet-3Blocks~\cite{huang2017deep},
Part-aware LSTM~\cite{shahroudy2016ntu},
ST-LSTM+Trust Gate\cite{liu2018skeleton},
CNN+LSTM2~\cite{nunez2018convolutional},
Two-Stream RNN~\cite{wang2017modeling},
STA-LSTM~\cite{song2017end},
GCA-LSTM (stepwise)~\cite{liu2017global},
Clips+CNN+MTLN~\cite{ke2017new},
View invariant~\cite{liu2017enhanced},
CNN+DPRL~\cite{tang2018deep},
VA-LSTM~\cite{zhang2017view},
ST-GCN~\cite{yan2018spatial},
CNN+LSTM~\cite{li2017skeleton},
Two-Stream CNN~\cite{du2015skeleton},
and SR-TSL~\cite{si2018skeleton}.
As we can see in Table~\ref{tab:NTU},
DACNN did not beat SR-TSL and CNN+LSTM in recognition accuracy.
Nevertheless, It still achieves a competitive result compared to other recent state-of-the-art algorithms,
such as ST-CGN, CNN+DRPL, and VA-LSTM, and even outperforms CNN+LSTM in the CS settings.

It is important to notify that, the NTU dataset is recorded in a very constrained experimental setting, 
which is an advantage for the methods which considerably rely on the spatial processing of human poses (such as SR-TSL and CNN+LSTM). 
\begin{table}[!t]
			\footnotesize		
	\centering	
	\begin{center}
		\begin{tabularx}{0.8\textwidth}{lXX|lXX} 
			\toprule				
			Method   & CS & CV&Method   & CS & CV \\
			\midrule		
			HBRNN-L~\cite{du2015hierarchical}
				&59.1&	64.0
				&Clips+CNN~\cite{ke2017new}&	79.6&	84.8\\
			Dynamic Sk.
			~\cite{hu2015jointly}&	60.2&	65.2&
			View invariant~\cite{liu2017enhanced}&	80.0&	87.2\\
			LieNet-3Blocks~\cite{huang2017deep}&	63.1&	68.4&
			CNN+DPRL~\cite{tang2018deep}&	82.3&	87.7\\
			Part-aware LSTM~\cite{shahroudy2016ntu}&	62.9&	70.3&
			VA-LSTM~\cite{zhang2017view}&	79.5&	87.9\\
			CNN+LSTM2~\cite{nunez2018convolutional}&	67.5&	76.2&
			ST-GCN~\cite{yan2018spatial}&	81.5&	88.3\\
			ST-LSTM+TG~\cite{liu2018skeleton}&	69.2&	78.7
			&Two-Stream CNN~\cite{du2015skeleton}&	83.1&	89.1\\
			Two-Stream RNN~\cite{wang2017modeling}&	72.1&	79.7&
			CNN+LSTM~\cite{li2017skeleton}&	82.9&	91.0\\
			STA-LSTM~\cite{song2017end}&	74.1&	81.8
			&SR-TSL~\cite{si2018skeleton}&	\textbf{84.8}&	\textbf{92.4}\\
			GCA-LSTM ~\cite{liu2017global}& 76.3&	84.5&
			DACNN (\textbf{Proposed})	&	83.8 &90.7\\
			\bottomrule			
		\end{tabularx}
\end{center}
\caption{Recognition accuracy ($\%$) for NTU dataset regarding Cross-View (CV) and Cross-Subject (CS).}	
\label{tab:NTU}
\end{table}

\subsection{Interpreting the Abs-filters}\label{sec:vis}
\begin{figure}[!b]
	\begin{center}
		\includegraphics[width=0.8\linewidth]{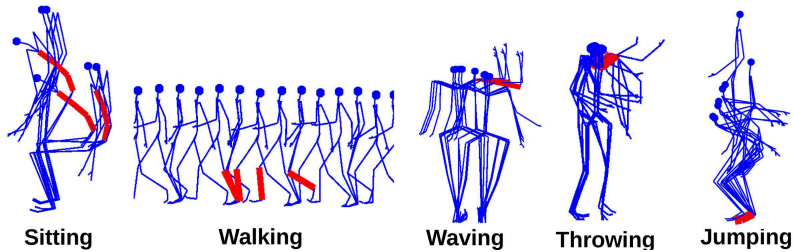}
	\end{center}
	\caption{Visualization of the Abs-filters learned by DACNN on the NTU dataset and the classes to which they are mostly related. Red links indicate the body parts in which the Abs-filters have high alignment values.}
	\label{fig:prots}
\end{figure}

%
%
%
Apart from the action recognition and segmentation performance, it is interesting to visualize the learned Abs-filters of DACNN in order to investigate any semantic pattern among them.
Relevantly, another strength of DACNN compared to other deep neural networks is its simplicity in visualization and interpretation of its trained filters (Abs-filters). 
Specifically, we associate filter $\nff{p}_i$ to class $c$ if 
\begin{equation}
c=\underset{c}{\argmax} \underset{\nY \in \text{class~} c}{\sum} \|\vec{v}^p(i,:)|_{\nY}\|_2^2,
\end{equation} 
in which $\vec{v}^p(i,:)|_{\nY}$ denotes the $i$-th row of the alignment map $\MB{V}^p$ after applying $\nff{p}_i$ on the dimensions of $\nY$.
In Fig.~\ref{fig:prots}, we visualize some of the Abs-filters learned by DACNN after being trained on the NTU dataset.
These filters are mostly related to the action classes \textit{walking}, \textit{waving}, \textit{sitting}, \textit{jumping}, and \textit{throwing}.
It is clear that each filter has learned a semantic subsequence from one joint of the whole action.
For instance, one Abs-filter is aligned with the movement of the \textit{left foot} in the \textit{jumping} action before the subject leaves the ground, or another filter has learned half of a \textit{walking} cycle on the \textit{right foot}.
Considering other Abs-filters which are illustrated in Fig.~\ref{fig:prots}, each of them has recognized a specific temporal pattern which facilitates the separation of that class from others.
\subsection{Ablation Study}\label{sec:abl}
To study individual roles of the different modules in DACNN (Fig.~\ref{fig:alg}), we perform an ablation study by 
repeating the recognition experiments (Sec.~\ref{sec:act}) for the variants of DACNN framework (Table~\ref{tab:abl}).
%
In DACNN-nf, we remove the fine-prediction module, 
and we do not employ any Abs-filter in DACNN-Al.
The DACNN-1D network is similar to DACNN-Al, but it uses 1D-conv filters instead of the Al-filters.
According to the results in Table~\ref{tab:abl}, 
%
the lower performance of DACNN-nf compared to DACNN shows the positive effect of the Abs-filter skip-connections (Fig.~\ref{fig:fine}) in improving the accuracy of the final prediction in 1D-CNN.

Nevertheless, the accuracy of DACNN-nf is still close to DACNN and is even higher than the state-of-the-art for SYS-3D and Montalbano. 
Removing the Abs-filters from DACNN causes a notable decrease in the performance of DACNN-Al, which justifies the significant role of these filters in extracting the key patterns in the data.
However, DACNN-Al obtains higher accuracies than DACNN-1D which proves how effective the alignment filters are regarding the SBARS problems.
%
\begin{table}[t]
	\footnotesize		
	\centering	
	\begin{center}
		\begin{tabularx}{0.6\textwidth}{Xcccc} %
			\toprule
			Method   & SYS-3D & NTU CS& NTU CV & Mont.\\
			\midrule		
			DACNN-1D &71.8	&70.5	&75.5&	70.54\\
			DACNN-Al & 78.4&	77.4&	85.3&	82.53\\
			DACNN-nf &82.5	&81.8	&88.9	&84.15\\
			\midrule
			DACNN (Fig.~\ref{fig:alg})   &\textbf{84.3}	&\textbf{83.0}	&\textbf{90.7}	&\textbf{85.2}\\
			\bottomrule
		\end{tabularx}
\end{center}
\caption{Prediction accuracies ($\%$) for partial implementations of DACNN on three selected datasets.}
\label{tab:abl}
\end{table}
\section{Conclusions}
In this work, we proposed a deep-aligned convolutional neural network for skeleton-based action recognition and temporal segmentation.
%
This network is constructed upon introducing the novel concept of temporal alignment filters for CNNs.
These filters are efficient choices for classification of skeleton-based motion data compared to regular convolution filters.
They extract crucial local patterns in the temporal dimensions of the data to better discriminate the action classes.
Besides the competitive performance of our DACNN framework compared to the state-of-the-art,  
its extracted features (learned Abs-filters) are easily interpretable
regarding their semantic contents.
%
%
Our empirical evaluation of DCANN on different SBARS benchmarks support our claims regarding the performance and benefits of our network.
We believe that the idea of incorporating alignments in CNNs can be further studied in other relevant areas such as
relevance analysis and generative adversarial networks.

\section*{Acknowledgement}
This research was supported by the Center of Cognitive 
Interaction Technology 'CITEC' (EXC 277) at Bielefeld University, which
is funded by the German Research Foundation (DFG).
\bibliographystyle{unsrt}
\bibliography{c:/Thesis/Publications/Ref4Papers_CS}



\end{document}